\documentclass{article}

\usepackage[nonatbib,final]{neurips_2019}

\usepackage[utf8]{inputenc} %
\usepackage[T1]{fontenc}    %
\usepackage{hyperref}       %
\usepackage{url}            %
\usepackage{booktabs}       %
\usepackage{amsfonts}       %
\usepackage{nicefrac}       %
\usepackage{microtype}      %
\usepackage{cite}
\usepackage{graphicx}
\usepackage{subfigure}
\usepackage{amssymb, amsmath}
\usepackage{xcolor}
\usepackage{stmaryrd}
\usepackage{wrapfig}
\usepackage{blindtext}
\usepackage{lipsum}

\title{Meta-Reinforced Synthetic Data for One-Shot Fine-Grained Visual Recognition}
\author{%
Satoshi Tsutsui\\
Indiana University\\
USA \\
\texttt{stsutsui@indiana.edu} \\
\And
Yanwei Fu\thanks{Y. Fu is with School of Data Science, and
MOE Frontiers Center for Brain Science,  Shanghai Key Lab of Intelligent Information Processing, Fudan University. } \\
Fudan University \\ China \\
\texttt{yanweifu@fudan.edu.cn} \\
\And
David Crandall \\
Indiana University\\
USA \\
\texttt{djcran@indiana.edu} \\
}

\begin{document}

\maketitle

\begin{abstract}
One-shot fine-grained visual recognition often
suffers from the problem of training data scarcity for new fine-grained
classes. To alleviate this problem, an off-the-shelf image generator
can be applied to synthesize additional training images, but
these synthesized images are often  not helpful for actually improving the accuracy of one-shot fine-grained
recognition.  This paper proposes a meta-learning framework to
combine generated images with original images, so that the resulting ``hybrid'' training images
can improve one-shot learning. Specifically, the generic image
generator is updated by a few training instances of novel classes, and a
Meta Image Reinforcing Network (MetaIRNet) is proposed to conduct
one-shot fine-grained recognition as well as image reinforcement. The
model is trained in an end-to-end manner, and our experiments
demonstrate consistent improvement over baselines on one-shot
fine-grained image classification benchmarks.
\end{abstract}

\section{Introduction}
The availability of vast labeled datasets has been crucial for the
recent success of deep learning. However, there will always be
learning tasks for which labeled data is sparse.  Fine-grained visual
recognition is one typical example: when images are to be
classified into many very specific categories (such as
species of birds), it may be difficult to obtain training examples for
rare classes, and producing the ground truth labels may require
significant expertise (e.g., ornithologists).
  One-shot learning is thus very desirable for fine-grained visual recognition.

A recent approach to address data scarcity is
meta-learning~\cite{yuxiong2016eccv,santoro2016meta,finn2017model,img_deform_2019},
which trains a parameterized function called a meta-learner that maps
labeled training sets to classifiers. The meta-learner is trained by
sampling small training and test sets from a large dataset of a base
class. Such a meta-learned model can be adapted to recognize novel
categories with a single training instance per class.  Another way to
address data scarcity is to synthesize additional training examples,
for example by using off-the-shelf Generative Adversarial Networks
(GANs)~\cite{goodfellow2014generative,biggan}.  However, classifiers
trained from GAN-generated images are typically inferior to those
trained with real images, possibly because the distribution of
generated images may be biased towards frequent patterns ({modes}) of the original image distribution
\cite{shmelkov2018good}. This is especially true in one-shot
{fine-grained} recognition where a tiny difference
(e.g., beak of a bird) can make a large difference in class.

To address these issues, we develop an approach to apply 
off-the-shelf generative models to synthesize training data in a
way that improves one-shot fine-grained classifiers. We begin by 
conducting a pilot study to transfer a
generator pre-trained on ImageNet in a one-shot scenario. We  show
that the generated images can indeed improve the performance of a
one-shot classifier when used with a carefully-designed rule to
combine the generated images with the originals. Based on these
preliminary results, we propose a  meta-learning approach to
learn these rules to reinforce the generated images effectively for
few-shot classification.

Our approach has two steps. First, an off-the-shelf generator trained
from ImageNet is updated towards the domain of novel classes by using
only a single image (Sec.~\ref{sec:method-gen}). Second,  since previous
work and our pilot study (Sec. \ref{sec:pilot-study}) suggest that
simply adding synthesized images to the training data may not improve
one-shot learning, the synthesized images are ``mixed'' with the
original images in order to bridge the domain gap between the two
(Sec.~\ref{sec:method-mix}). The effective mixing strategy is learned
by a meta-learner, which essentially
 boosts the performance of fine-grained categorization with
a single training instance per class. Lastly, we experimentally validate that our
approach can achieve improved performance over  baselines on
fine-grained classification datasets in one-shot situations
(Sec.~\ref{sec:experiment}).

To summarize, the contributions of this paper are: (1) a method to
transfer a pre-trained generator with a single image, (2) a method to
learn to complement real images with synthetic images in a way that
benefits one-shot classifiers, and (3) to experimentally demonstrate
that these methods improve one-shot classification accuracy on
fine-grained visual recognition benchmarks.

\section{Related Work}\label{sec:related}
\noindent \textbf{Image Generation.}  Learning to generate realistic
images has many potential applications, but is challenging with
typical supervised learning. Supervised learning minimizes a loss
function between the predicted output and the desired output but, for
image generation, it is not easy to design such a
perceptually-meaningful loss between images.  Generative Adversarial
Networks (GANs)\cite{goodfellow2014generative} address this issue by
learning not only a generator but also a loss function --- the
discriminator --- that helps the generator to synthesize images
indistinguishable from real ones.  
This
adversarial learning is intuitive but is known to often be
unstable~\cite{gulrajani2017improved} in practice. Recent progress
includes better CNN
architectures~\cite{radford2015unsupervised,biggan}, training
stabilization
tips~\cite{arjovsky2017wasserstein,gulrajani2017improved,miyato2018spectral},
and interesting applications (e.g.~\cite{CycleGAN2017}). In
particular, BigGAN~\cite{biggan} trained on ImageNet has shown
visually impressive generated images with stable performance on
generic image generation tasks.  Several
studies~\cite{noguchi2019image,wang2018transferring} have explored
generating images from few examples, but their focus has not been on
one shot classification.  Several
papers~\cite{de2017modulating,dumoulin2017learned,noguchi2019image}
use the idea of adjusting batch normalization layers, which helped
inspire our work. Finally, some work has investigated using GANs to
help image
classification~\cite{shmelkov2018good,shrivastava2017learning,antoniou2018augmenting,zhang2018metagan,gao2018low};
our work differs in that we apply an off-the-shelf generator
pre-trained from a large and generic dataset.

\noindent \textbf{Few-shot Meta-learning.} Few shot
classification~\cite{chen2019closer} is a sub-field of meta-learning
(or ``learning-to-learn'') problems, in which the task is to train a
classifier with only a few examples per class.  Unlike the typical
classification setup, in few-shot classification the labels in the
training and test sets have no overlapping categories. Moreover, the
model is trained and evaluated by sampling many few-shot tasks (or
episodes). For example, when training a dog breed classifier, an
episode might train to recognize five dog species with only a single training
image per class --- a 5-way-1-shot setting.  A meta-learning method
trains a meta-model by sampling many episodes from training classes
and is evaluated by sampling many episodes from other unseen
classes. With this episodic training, we can choose several possible
approaches to learn to learn. For example, ``learning to compare''
methods learn a metric space
(e.g.,\cite{vinyals2016matching,snell2017prototypical,sung2018learning}),
while other approaches learn to fine-tune (e.g.,
\cite{finn2017model,finn2018probabilistic,rusu2018meta,ravi2017optimization})
or learn to augment data (e.g.,
\cite{wang2018low,schwartz2018delta,hariharan2017low,chen2019imageblock_aaai,gao2018low}). An
advantage of the latter type is that, since it is data augmentation,
we can use it in combination with any other approaches. Our approach
also explores data augmentation by mixing the original images with
synthesized images produced by a fine-tuned generator, but we find
that the naive approach of simply adding GAN generated images to the
training dataset does not improve performance.  But by carefully
combining generated images with the original images, we find that
we can effectively synthesize examples that contribute to increasing
the performance. Thus meta-learning is employed to learn the proper
combination strategy.

\section{Pilot Study}
\label{sec:pilot-study} 

To explain how we arrived at our approach, we describe some initial
experimentation which motivated the development of our methods.

\begin{table}
	\caption{CUB  5-way-1-shot classification accuracy (\%)
          using ImageNet features. Simply adding generated images to the training
          set does not help, but adding hybrid images, as
          in Fig. \ref{fig:fintune-gan-samples} (h), can.}
	\label{tbl:pilot-study}
	\centering
	\begin{tabular}{l@{\,\,\,\,\,\,\,\,\,\,}cccc}
		\toprule
		Training Data      &  Nearest Neighbor  & Logistic Regression &  Softmax Regression  \\
		\midrule
		Original                            & 69.6 &  75.0  &  74.1  \\
		Original + Generated        & 70.1  &  74.6  &  73.8  \\
		Original + Mixed               & 70.6  & 75.5  & 74.8    \\
		\bottomrule
	\end{tabular}
\end{table}
\begin{figure}
	\centering
	\vspace{-1mm}
	\begin{subfigure}[]{%
			\includegraphics[clip, width=0.1\columnwidth]{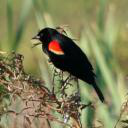}}
	\end{subfigure}
	\vspace{-1mm}
	\begin{subfigure}[]{%
			\includegraphics[clip, width=0.1\columnwidth]{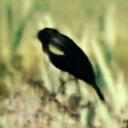}}
	\end{subfigure}
	\begin{subfigure}[]{%
			\includegraphics[clip, width=0.1\columnwidth]{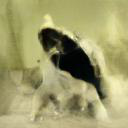}}
	\end{subfigure}
	\begin{subfigure}[]{%
			\includegraphics[clip, width=0.1\columnwidth]{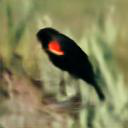}}
	\end{subfigure}
	\begin{subfigure}[]{%
			\includegraphics[clip, width=0.1\columnwidth]{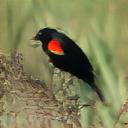}}
	\end{subfigure}
	\begin{subfigure}[ ]{%
	\includegraphics[clip, width=0.1\columnwidth]{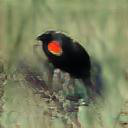}}
	\end{subfigure}
	\begin{subfigure}[]{%
	\includegraphics[clip, width=0.1\columnwidth]{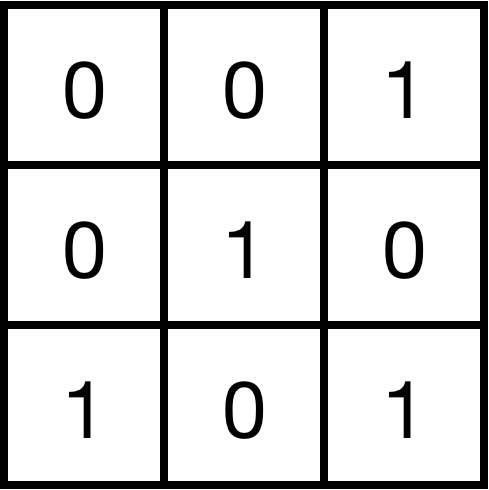}}
	\end{subfigure}
	\begin{subfigure}[ ]{%
	\includegraphics[clip, width=0.1\columnwidth]{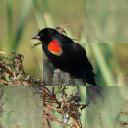}}
	\end{subfigure}
	\caption{Samples described in Sec. \ref{sec:pilot-study}. (a)
          Original image. (b) Result of tuning noise only. (c) Result of tuning the whole
          network. (d) Result of tuning batch norm only. (e) Result of tuning batch norm with
          perceptual loss. (f) Result of slightly disturbing noise from (e). (g) a
          $3 \times 3$ block weight matrix $w$. (g) Result of mixing (a) and (f) as
          $w \times$(f) + $(1 - w )\times$(a). 
        }\label{fig:fintune-gan-samples}
\end{figure}
\paragraph{How can we  transfer generative knowledge from
pre-trained GANs?}  We aim to quickly generate training images for
few-shot classification. Performing adversarial learning
(\emph{i.e}. training generator and discriminator initializing with
pre-trained weights) is not practical when we only have one or two
examples per class. Instead, we want to develop a method that does not
depend on the number of images at all; in fact, we consider the
extreme case where only a single image is available, and want to
generate variants of the image using a pre-trained GAN. We tried fixing
the generator weights and optimizing the noise so that it generates
the target image, under the assumption that sightly modifying the
optimized noise would produce a variant of the original. However,
naively implementing this idea with BigGAN did not reconstruct the
image well, as shown in the sample in
Fig. \ref{fig:fintune-gan-samples}(b). We then tried fine-tuning the generator
weights also, but this produced even worse images stuck in a local
minima, as shown in Fig. \ref{fig:fintune-gan-samples}(c).  

We speculate that the best approach may be somewhere between the two
extremes of tuning noise only and tuning both noise and
weights. Inspired by previous
work~\cite{de2017modulating,dumoulin2017learned,noguchi2019image}, we
propose to fine-tune only scale and shift parameters in the batch
normalization layers. This strategy produces better images as shown in
Fig. \ref{fig:fintune-gan-samples}(d). Finally, again inspired by
previous work\cite{noguchi2019image}, we not only minimize the
pixel-level distance but also the distance of a pre-trained CNN
representation (i.e. perceptual loss~\cite{johnson2016perceptual}),
and we show the slightly improved results in Fig.
\ref{fig:fintune-gan-samples}(e). We can also generate slightly
different versions by adding random perturbations to the tuned noise
(e.g., the ``fattened'' version of the same bird in
Fig. \ref{fig:fintune-gan-samples}(f)).  The entire training process
needs fewer than 500 iterations and takes less than 20 seconds on an
NVidia Titan Xp GPU. We explain the resulting generation strategy developed based on this pilot study in
Sec. \ref{sec:method}.

\paragraph{Are generated images helpful for few shot learning?}
Our goal is not to generate images, but to augment the training data
for few shot learning. A naive way to do this is to apply the above generation
technique for each training image, in order to double the training set. We
tested this idea on a validation set (split the same as \cite{chen2019closer}) from
the Caltech-UCSD bird dataset~\cite{WahCUB_200_2011} and computed average accuracy on 100 episodes of
5-way-1-shot classification. We used pre-trained ImageNet features from
ResNet18~\cite{he2016deep} with nearest neighbor, one-vs-all logistic
regression, and softmax regression (or multi-class logistic regression).
As shown in Table~\ref{tbl:pilot-study}, the accuracy actually drops for two of the three classifiers when we double
the size of our training set by generating synthetic training images, suggesting
that the
generated images are harmful for training classifiers. 

\noindent \textbf{What is the proper way of synthesizing images to
  help few-shot learning?} Given that the synthetic images
\textit{appear} meaningful to humans, we conjecture that they can
benefit few shot classification when properly mixed with originals to create hybrid images. To empirically test this hypothesis, we devised a random $3\times3$
grid to combine the images. As shown in
Fig. \ref{fig:fintune-gan-samples}(h), images (a) and (f) were combined by
taking a linear combination within each cell of the grid of (g).
Finally, we added mixed images like (h) into the training data, and
discovered that this produced a modest increase in accuracy (last row
of Table \ref{tbl:pilot-study}). While the increase is marginal, these
mixing weights were binary and manually selected, and thus likely not
optimal.  In Sec.~\ref{sec:method-mix}, we show how to learn this
mixing strategy in an end-to-end manner using a meta-learning
framework.

\begin{figure}
  \centering
    \includegraphics[width=0.9\columnwidth]{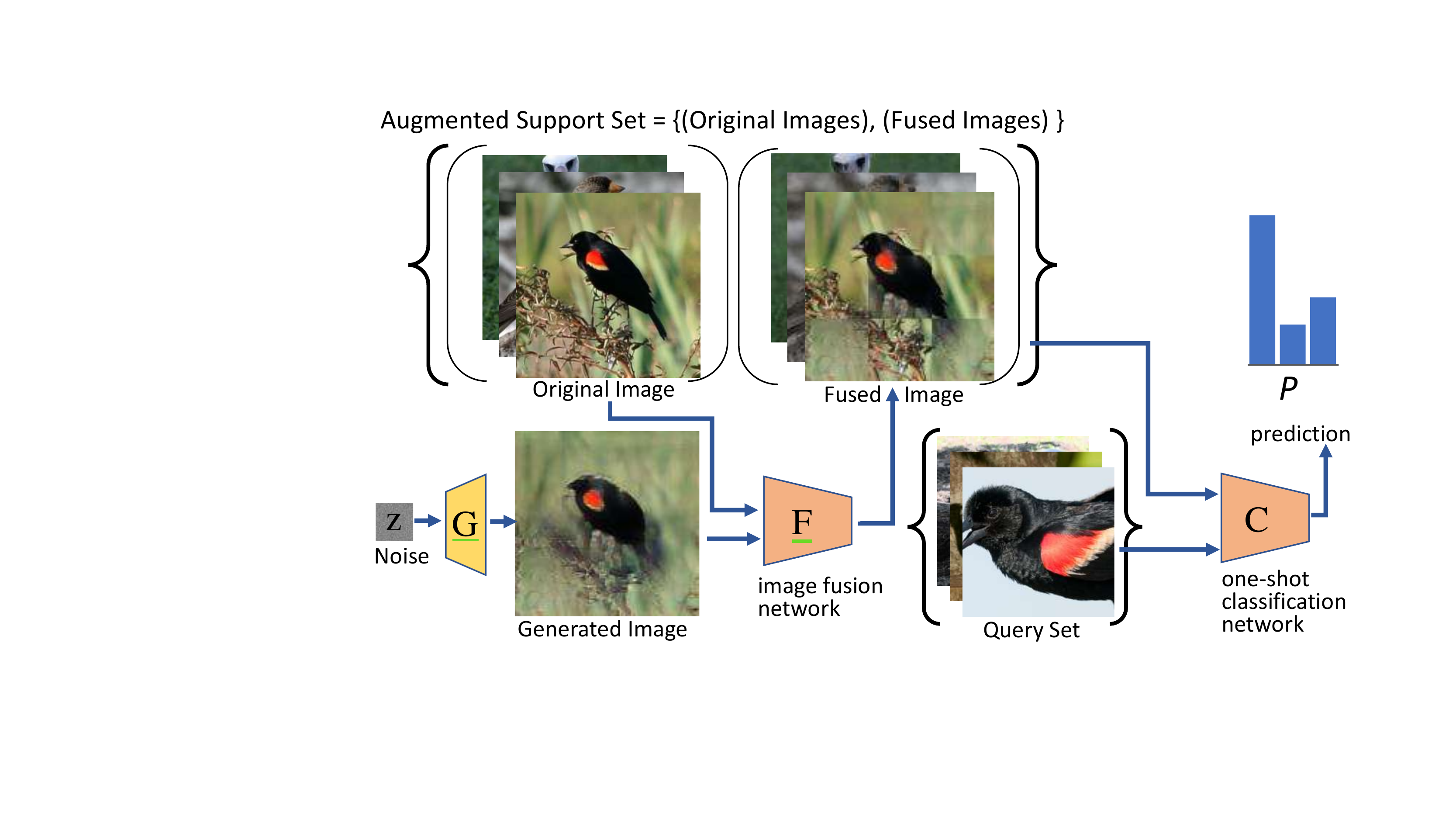}
    \caption{\textbf{Our Meta Image Reinforcing Network (MetaIRNet) } has
      two modules: an image fusion network, and a one-shot classification
      network. The image fusion network reinforces generated
      images to try to make them beneficial for the one-shot classifier, while the
      one-shot classifier learns representations that are suitable to
      classify unseen examples with few examples. Both networks
      are trained  end-to-end, so the loss back-propagates from
      classifier to the fusion network.}\label{fig:method}
\end{figure}

\section{Method}
\label{sec:method}

The results of the pilot study in the last section suggested that 
producing synthetic images could be useful for few-shot fine-grained
recognition, but only if it is done in a careful way. In this section,
we use these findings to propose a novel technique for doing this
effectively.
We
propose a GAN fine-tuning method that works with a single image
(Sec.~\ref{sec:method-gen}), and an effective meta-learning method to not
only learn to classify with few  examples, but also to learn to
effectively reinforce the generated images
(Sec.~\ref{sec:method-mix}).

\subsection{Fine-tuning Pre-trained Generator for Target Images}
\label{sec:method-gen} 

GANs typically have a generator $G$ and a discriminator $D$. Given an input
signal $z\sim\mathcal{N}(0,1)$, a well-trained generator synthesizes
an image $G(z)$. In our tasks, we adapt an off-the-shelf GAN generator
$G$ that is pre-trained on the ImageNet-2012 dataset in order to generate
more images in a target, data-scarce domain.  Note that we do not use the
discriminator, since adversarial training with a few images is
unstable and may lead to model collapse. Formally, we fine-tune $z$ and
the generator $G$   such that $G$ generates an image $\mathbf{I}_{z}$
from an input vector $z$ by minimizing the distance between $G(z)$ and
$\mathbf{I}_{z}$, where the vector $z$ is randomly
initialized. Inspired by previous
work~\cite{chen2016infogan,arjovsky2017wasserstein,noguchi2019image},
we minimize a loss function $\mathcal{L}_{G}$ with 
$\mathcal{L}_{1}$ distance and perceptual loss $\mathcal{L}_{perc}$
with earth mover regularization $\mathrm{\mathcal{L}}_{EM}$,
\begin{align}
\mathcal{L}_{G}\left(G,\mathbf{I}_{z},z\right)=\mathcal{L}_{1}\left(G(z),\mathbf{I}_{z}\right)+\lambda_{p}\mathcal{L}_{perc}\left(G(z),\mathbf{I}_{z}\right)+\lambda_{z}\mathrm{\mathcal{L}}_{EM}\left(z,r\right),\label{eq:loss-gen}
\end{align}
 where  $\mathrm{\mathcal{L}}_{EM}$ is an earth mover distance between
$z$ and random noise $r\sim\mathcal{N}(0,1)$ to regularize $z$ to
be sampled from a Gaussian, and $\lambda_{p}$ and $\lambda_{z}$
are coefficients of each term.

Since only a few training images are available in the target domain, only
scale and shift parameters of the batch normalization of $G$ are updated
in practice. Specifically, only the $\gamma$ and $\beta$
of each batch normalization layer are updated in each layer,
\begin{equation}
\hat{x}=\frac{x-\mathrm{\mathbb{E}}(x)}{\sqrt{\mathrm{Var}(x)+\epsilon}}\qquad h=\gamma\hat{x}+\beta,\label{eq:scale=000026gamma}
\end{equation}
\noindent where $x$ is the input feature from the previous layer, and $\mathbb{E}$
and $\mathrm{Var}$ indicate the mean and variance functions, respectively. Intuitively
and in principle, updating $\gamma$ and $\beta$ only is equivalent
to adjusting the activation of each neuron in a layer.  Once updated,
the $G(z)$ would be synthesized to reconstruct the image $\mathbf{I}_{z}$.
Empirically, a small random perturbation $\epsilon$ is added to
$z$ as $G\left(z+\epsilon\right)$.
Examples of $\mathbf{I}_{z}$, $G(z)$ and $G\left(z+\epsilon\right)$
are illustrated in in Fig. \ref{fig:fintune-gan-samples} (a), (e),
and (f), respectively.

\subsection{Meta Reinforced Synthetic Data for Few-shot Learning \label{sec:method-mix}}
We propose a meta-learning method to add synthetic data to the originals.

\noindent \textbf{One-shot Learning.} One-shot classification is a
meta-learning problem that divide a dataset into two sets:
meta-training (or base) set and meta-testing (or novel) set. The
classes in the base set and the novel sets are disjoint. In other
words, we have $\mathcal{D}_{base}=\left\{
\left(\mathbf{I}_{i},y_{i}\right), y_{i}\in\mathcal{C}_{base}\right\}
$ and $\mathcal{D}_{novel}=\left\{ \left(\mathbf{I}_{i},y_{i}\right),
y_{i}\in\mathcal{C}_{novel}\right\}$ where $\mathcal{C}_{base}\cup
\mathcal{C}_{novel}=\emptyset$.  The task is to train a classifier on
$\mathcal{D}_{base}$ that can quickly generalize to unseen classes in
$\mathcal{C}_{novel}$ with one or few examples.  To do this, a
meta-learning algorithm performs meta-training by sampling many
one-shot tasks from $\mathcal{D}_{base}$, and is evaluated by sampling
many similar tasks from $\mathcal{D}_{novel}$. Each sampled task
(called an episode) is an $n$-way-$m$-shot classification problem with
$q$ queries, meaning that we sample $n$ classes with $m$ training and
$q$ test examples for each class.  In other words, an episode has a
support (or training) set $S$ and a query (or test) set $Q$, where
$|S|=n \times m$ and $|Q|=n\times q$. One-shot learning means $m=1$. The
notation $S_c$ means the support examples only belong to the class
$c$, so $|S_c| = m$.

\noindent \textbf{Meta Image Reinforcing Network (MetaIRNet).} We
propose a Meta Image Reinforcing Network (MetaIRNet), which not only
learns a few-shot classifier, but also learns to reinforce generated
images by combining real  and 
generated images. MetaIRNet is composed of two modules: an image fusion
network $F$, and a one-shot
classification network $C$.

\noindent \textit{The Image Fusion Network}
$F$ combines a real image
$\mathbf{I}$ and a corresponding generated image
$\mathbf{I}_{g}$ into a new image $\mathbf{I}_{syn}=F\left(\mathbf{I},\mathbf{I}_{g}\right)$ that is beneficial
for training a one-shot classifier. 
Among the many possible ways to synthesize images, 
we were inspired by a 
block augmentation method~\cite{chen2019imageblock_aaai} and
use  
grid-based linear combination. As shown in
Figure~\ref{fig:fintune-gan-samples}(g), we divide the images into a
$3\times3$ grid and linearly combine the cells with the weights
$\mathbf{w}$ produced by a CNN conditioned on the two images. That is,
\begin{equation}
\mathbf{I}_{syn}=\mathbf{w} \odot \mathbf{I}+\left(1-\mathbf{w}\right) \odot \mathbf{I}_{g}
\end{equation}
where $\odot$ is element-wise multiplication, and $\mathbf{w}$ is
resized to the image size keeping the block structure. The CNN to produce $\mathbf{w}$ extracts the feature vectors of $\mathbf{I}$ and
$\mathbf{I}_g$, concatenates them, and uses a fully-connected
layer to produce a weight corresponding to each of the nine cells in the $3
\times 3$ grid.  Finally, for each real image
$\mathbf{I}^{i}$, we generate $n_{aug}$ images, producing $n_{aug}$
synthetic images, and assign the same class label $y_{i}$ to 
each synthesized image $\mathbf{I}_{syn}^{i,j}$ to obtain an augmented
support set,
\begin{equation}
\tilde{S}=\left\{ \left( \mathbf{I}^{i}_{},y^{i}_{}\right),\left\{ \left(\mathbf{I}_{syn}^{i,j},y_{}^{i}\right)\right\}_{j=1}^{n_{aug}}\right\}_{i=1}^{n\times m} .
\end{equation}

\noindent The \textit{One-Shot Classification Network}
$C$ maps an input image $\mathbf{I}$ into
feature maps $C\left(\mathbf{I}\right)$, and performs
one-shot classification. Although any one-shot classifier can be
used, we choose the non-parametric prototype classifier
of Snell \textit{et al.}~\cite{snell2017prototypical} due to its superior performance and
simplicity. During each episode, given the sampled $S$ and $Q$,
the image fusion network produces an augmented support set
$\tilde{S}$. This classifier computes the prototype vector $\mathbf{p}_c$ for each class $c$ in $\tilde{S}$ as an average feature
vector, 
\begin{equation}
\mathbf{p}_c = \frac{1}{|\tilde{S_c}|}\sum_{(\mathbf{I}_{i},y_{i})\in\tilde{S_c}}C\left(\mathbf{I}_{i}\right). 
\end{equation}

For a query image $\mathbf{I}_{i}\in Q$,
the probability of belonging to a class $c$ is estimated as, 
\begin{equation}
P\left(y_{i}=c\text{\ensuremath{\mid}}\mathbf{I}_{i}\right)=
\frac{\mathrm{exp}\left(-\left\Vert  C\left(\mathbf{I}_{i}\right)-\mathbf{p}_c\right\Vert \right)}{\sum_{k=1}^{n}\mathrm{exp}\left(-\left\Vert C\left(\mathbf{I}_{i}\right)-\mathbf{p}_k\right\Vert \right)}\label{eq:prototypical_classifier}
\end{equation}
\noindent where $\parallel\cdot\parallel$ is the Euclidean
distance. Then, for a query image, the class with the highest
probability becomes the final prediction of the one-shot classifier.

\paragraph{Training}  In the meta-training phase, we jointly train  $F$ and $C$ in an  end-to-end manner, minimizing a cross-entropy loss function,
\begin{equation}
\min_{\mathbf{\theta}_F,\mathbf{\theta}_C}
\frac{1}{|Q|}\sum_{\left(\mathbf{I}_{i},y_{i}\right)\in
  Q}-\mathrm{log}P\left(y_{i}\mid\mathbf{I}_{i}\right),\label{eq:loss_function}
\end{equation} 
where $\mathbf{\theta}_F$ and $\mathbf{\theta}_C$ are the learnable parameters of $F$ and $C$.

\section{Experiments}\label{sec:experiment}
To investigate the effectiveness of our approach, we perform 1-shot-5-way
classification  following the meta-learning experimental
setup described in Sec. \ref{sec:method-mix}.  We perform 1000 episodes
in meta-testing, with 16 query images per class per episode, and
report average classification accuracy and 95\% confidence
intervals. We use the fine-grained classification dataset of Caltech UCSD
Birds (CUB)~\cite{WahCUB_200_2011} for our main experiments,
and another fine-grained dataset of North American Birds
(NAB)~\cite{van2015building} for secondary experiments. CUB has 11,788
images with 200 classes, and NAB has 48,527 images with 555 classes.

\subsection{Implementation Details}
 While our fine-tuning method introduced in Sec.~\ref{sec:method-gen}
 can generate images for each step in meta-training and meta-testing,
 it takes around 20 seconds per image, so we apply the generation
 method ahead of time to make our experiments more efficient.  We use
 a BigGAN pre-trained on ImageNet, using the publicly-available
 weights. We set $\lambda_p = 0.1$ and $\lambda_z = 0.1$, and perform
 500 gradient descent updates with the Adam~\cite{kingma2014adam}
 optimizer with learning rate $0.01$ for $z$ and $0.0005$ for the
 fully connected layers, to produce scale and shift parameters of the
 batch normalization layers.  We manually chose these hyper-parameters
 by trying random values from 0.1 to 0.0001 and visually checking the
 quality of a few generated images. We only train once for each image,
 generate 10 random images by perturbing $z$, and randomly use one of
 them for each episode ($n_{aug} = 1$). For image classification, we use
 ResNet18~\cite{he2016deep} pre-trained on ImageNet for the two CNNs
 in $F$ and one in $C$.  We train $F$ and $C$ with Adam with a default
 learning rate of $0.001$. We select the best model based on the
 validation accuracy, and then compute the final accuracy on the test
 set. For CUB, we use the same train/val/test split used in previous
 work~\cite{chen2019closer}, and for NAB we randomly split with a
 proportion of train:val:test = 2:1:1; see supplementary material for
 details. Further implementation details are available as supplemental
 source code.\footnote{http://vision.soic.indiana.edu/metairnet/}

\subsection{Comparative and Ablative Study on CUB dataset}
\paragraph{Baselines.} We compare our 
MetaIRNet with three types of baselines. (1) Non-meta learning
classifiers: We directly train the same ImageNet pre-trained CNN used
in $F$ to classify images in $\mathcal{D}_{base}$, and use it as a
feature extractor for $\mathcal{D}_{novel}$. We then use off-the-shelf
classifiers \textbf{nearest neighbor}, \textbf{logistic regression}
(one-vs-all classifier), and \textbf{softmax regression} (also called
multi-class logistic regression). (2) Meta-learning classifiers: We
try the meta-learning method of prototypical network
(\textbf{ProtoNet}~\cite{snell2017prototypical}).  ProtoNet computes
an average prototype vector for each class and performs nearest
neighbor with the prototypes. We note that our MetaIRNet adapts
ProtoNet as a choice of $F$ so this is an ablative version of our
model (MetaIRNet without the image fusion module). (3) Data
augmentation: Because our MetaIRNet learns data-augmentation as a
sub-module, we also compare with three data augmentation strategies,
Flip, Gaussian, and FinetuneGAN.  \textbf{Flip} horizontally flips the
images. \textbf{Gaussian} adds Gaussian noise with standard deviation
0.01 into the CNN features. \textbf{FinetuneGAN}~(introduced in
Sec. \ref{sec:method-gen}) generates augmented images by
fine-tuning the ImageNet-pretrained BigGAN with each support set. Note
that we do these augmentations in the meta-testing stage to increase
the support set.  For fair comparison, we use ProtoNet as the base
classifier of these data augmentation baselines.

\begin{table}
	\caption{5-way-1-shot accuracy (\%) on CUB/NAB dataset with ImageNet pre-trained ResNet18}
	\label{tbl:results-main}
	\centering
	\begin{tabular}{l l cc}
		\toprule
	    Method   & Data Augmentation   & CUB Acc. & NAB Acc. \\
		\midrule
		Nearest Neighbor    & - & $79.00\pm0.62$ &$80.58\pm0.59$\\
		Logistic Regression & - & $81.17\pm0.60$ &$82.70\pm0.57$\\
		Softmax Regression  & - & $80.77\pm0.60$ &$82.38\pm0.57$\\
		\midrule
		ProtoNet  & -      & $81.73\pm0.63$  & $87.91\pm0.52$ \\ 
		ProtoNet  & FinetuneGAN & $79.40\pm0.69$  & $85.40\pm0.59$ \\ 
		ProtoNet  & Flip   & $82.66\pm0.61$  & $88.55\pm0.50$ \\ 
		ProtoNet  & Gaussian &$81.75\pm0.63$ & $87.90\pm0.52$\\ 
        \midrule
        MetaIRNet (Ours)  & FinetuneGAN  & $84.13\pm0.58$ & $89.19\pm0.51$ \\
        MetaIRNet (Ours)  & FinetuneGAN, Flip & $\mathbf{84.80\pm0.56}$ & $\mathbf{89.57\pm0.49}$ \\
		\bottomrule
	\end{tabular}
\end{table}

\begin{table}
	\caption{5-way-1-shot accuracy (\%) on CUB dataset with Conv4 without ImageNet pre-training}
	\label{tbl:results-cub-scratch}
	\centering
	\begin{tabular}{l c c c c}
		\toprule
	    MetaIRNet & ProtoNet\cite{snell2017prototypical} &    MatchingNet~\cite{vinyals2016matching} & MAML~\cite{finn2017model}  &  RelationNet~\cite{sung2018learning}\\
		\midrule
		$\mathbf{65.86\pm0.72}$ & $63.50\pm0.70	$& $61.16\pm0.89$~\cite{chen2019closer} & $55.92\pm0.95$~\cite{chen2019closer} & $62.45\pm0.98$~\cite{chen2019closer} \\
		\bottomrule
	\end{tabular}
\end{table}

\paragraph{Results.}
As shown in Table~\ref{tbl:results-main}, our MetaIRNet is superior to
all baselines including the meta-learning classifier of ProtoNet
(84.13\% vs. 81.73\%) on the CUB dataset. It is notable that while ProtoNet
has worse accuracy when simply using the generated images as
data augmentation, our method shows an accuracy increase from ProtoNet,
which is equivalent to MetaIRNet without the image fusion module. This
indicates that our image fusion module can effectively complement the
original images while removing harmful elements from generated ones.

Interestingly, horizontal flip augmentation yields nearly a 1\% accuracy
increase for ProtoNet. Because flipping augmentation cannot be learned
directly by our method, we conjectured that our method could also
benefit from it. The final line of the table shows an additional experiment
with our MetaIRNet
combined with random flip augmentation, showing an additional accuracy increase  from 84.13\% to 84.80\%. This suggests that our method
provides an improvement that is orthogonal to flip
augmentation.

\begin{wrapfigure}{r}{0.25\textwidth}
    \vspace{-5mm}
	\centering
	\includegraphics[width=0.25\textwidth]
	{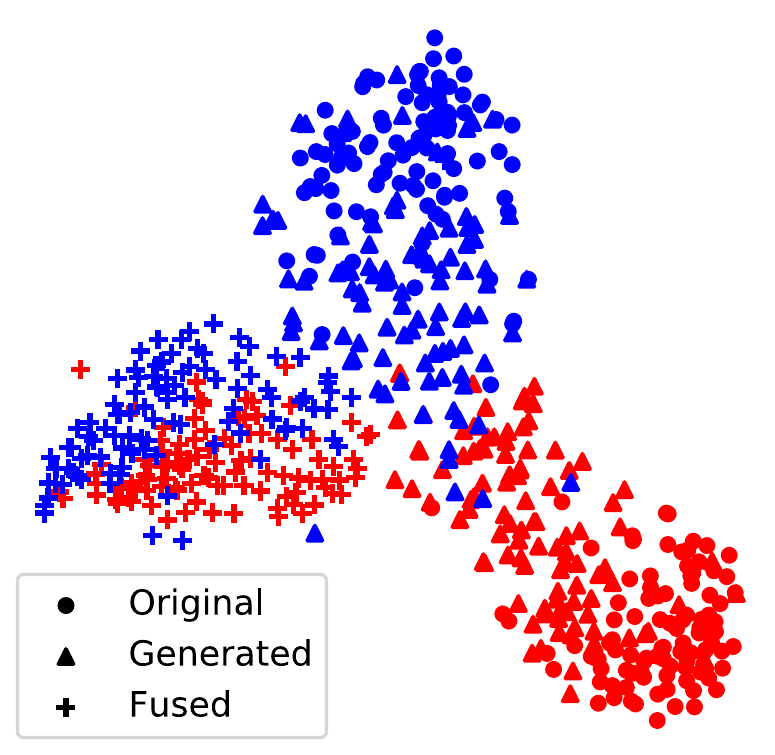}
	\caption{ t-SNE plot
	}\label{fig:tsne}
\end{wrapfigure}
\begin{figure}
\begin{tabular}{l}
	\hspace{-1.5mm}  Original \hspace{1.5mm}   Generated \hspace{3 mm}   Fused  \hspace{6.5mm}  Weight    
	\hspace{9mm}  Original \hspace{1.5mm}   Generated \hspace{3.5mm}   Fused  \hspace{6mm}  Weight    \\
\end{tabular} 

\includegraphics[width=0.11\columnwidth]
{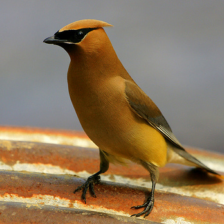}
\includegraphics[width=0.11\columnwidth]
{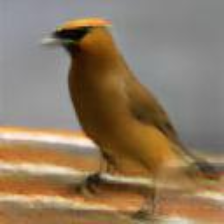}
\includegraphics[width=0.11\columnwidth]
{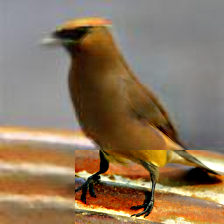}
\includegraphics[width=0.14\columnwidth]
{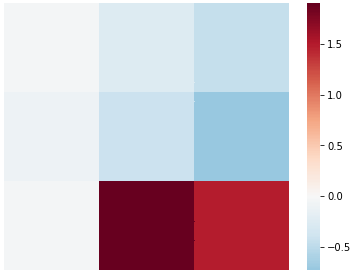}
\hspace{3mm}
\includegraphics[width=0.11\columnwidth]
{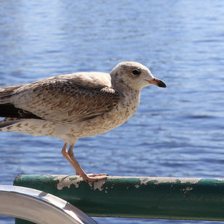} 
\includegraphics[width=0.11\columnwidth]
{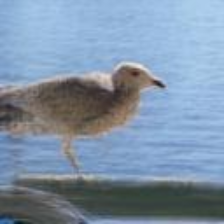}
\includegraphics[width=0.11\columnwidth]
{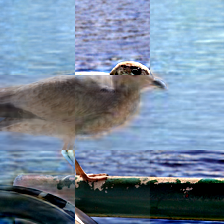}
\includegraphics[width=0.14\columnwidth]
{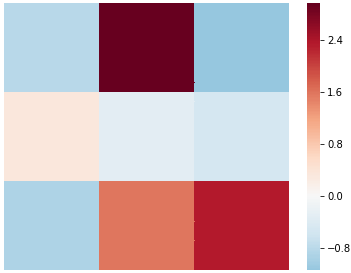} 

\includegraphics[width=0.11\columnwidth]
{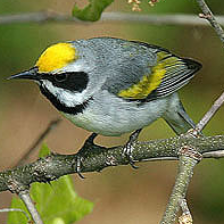}
\includegraphics[width=0.11\columnwidth]
{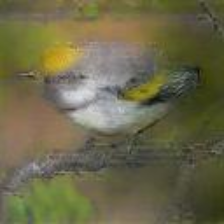}
\includegraphics[width=0.11\columnwidth]
{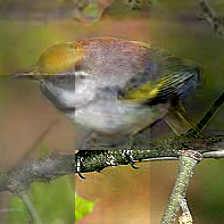}
\includegraphics[width=0.14\columnwidth]
{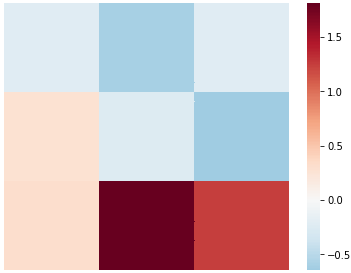}
\hspace{3mm}
\includegraphics[width=0.11\columnwidth]
{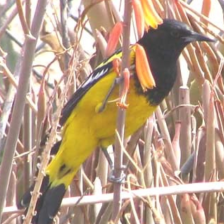} 
\includegraphics[width=0.11\columnwidth]
{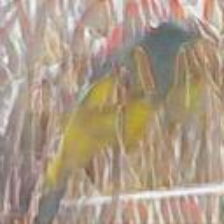}
\includegraphics[width=0.11\columnwidth]
{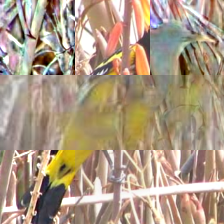}
\includegraphics[width=0.14\columnwidth]
{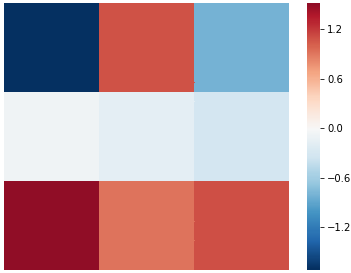} 

\includegraphics[width=0.11\columnwidth]
{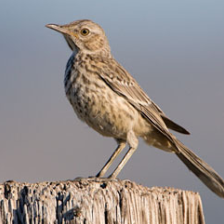}
\includegraphics[width=0.11\columnwidth]
{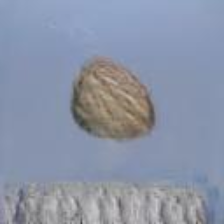}
\includegraphics[width=0.11\columnwidth]
{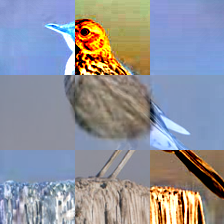}
\includegraphics[width=0.14\columnwidth]
{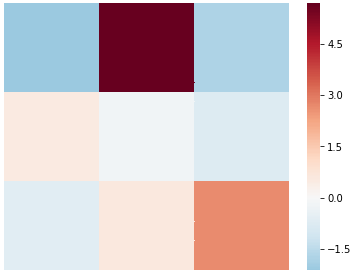}
\hspace{3mm}
\includegraphics[width=0.11\columnwidth]
{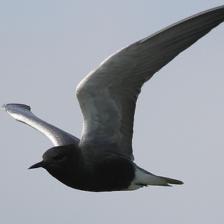} 
\includegraphics[width=0.11\columnwidth]
{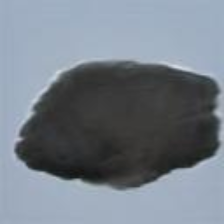}
\includegraphics[width=0.11\columnwidth]
{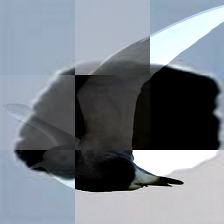}
\includegraphics[width=0.14\columnwidth]
{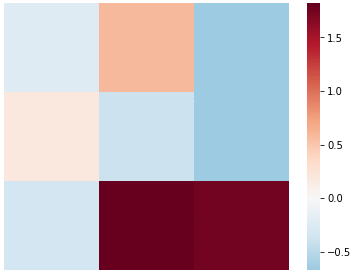} 

\caption{
	Samples of original image, generated image, fused image,  and mixing weight $\mathbf{w}$. Higher weight (red) means more original image used, and lower weight (blue) means more generated image used. We show three types of samples based on the quality of generated images: very good (top row), relatively good (middle row), and very bad or broken (last row).  }\label{fig:result-samples}
\end{figure}

\paragraph{Case Studies.}  
We show some sample visualizations in
Fig.~\ref{fig:result-samples}. We observe that image generation often
works well, but sometimes completely fails. An advantage of our
technique is that even in these failure cases, our fused images often
maintain some of the object's shape, even if the images themselves do
not look realistic.  In order to investigate the quality of generated
images in more detail, we randomly pick two classes,
sample 100 images for each class, and a show t-SNE visualization of real
images (\textbullet), generated images ($\blacktriangle$), and
augmented fused images (\textbf{+}) in Fig.  \ref{fig:tsne}, with classes shown in red and blue. It is
reasonable that the generated images are closer to the real ones,
because our loss function (equation 1) encourages this to be
so. Interestingly, perhaps due to artifacts of $3 \times 3$ patches,
the fused images are distinctive from the real/generated images,
extending the decision boundary.

\paragraph{Comparing with state-of-the-art meta-learning classifiers.}
It is a convention in the machine learning community to compare any
new technique with the performance of many state-of-the-art methods reported in the literature.
 This is somewhat difficult for us to do fairly, however:
we use ImageNet-pre-trained features as a starting point (which
is a natural design decision considering that our focus is how to use
ImageNet pre-trained generators for improving fine-grained one-shot
classification), but much of the one/few-shot learning literature
focuses on algorithmic improvements and thus trains from scratch
(often with non-fine-grained datasets). The Delta
Encoder~\cite{schwartz2018delta}, which uses the idea of learning data
augmentation in the feature space, reports  82.2\% on one-shot
classification on the CUB dataset with ImageNet-pre-trained features, but
this is an average of only 10 episodes. 

To provide more stable comparison, we cite a benchmark
study~\cite{chen2019closer} reporting accuracy of other
meta-learners~\cite{vinyals2016matching,finn2017model,sung2018learning}
on the CUB dataset with 600 episodes. To compare with these scores, we
experimented with our MetaIRNet and the ProtoNet baseline using the
same four-layered CNN.  As shown in Table
\ref{tbl:results-cub-scratch}, our MetaIRNet performs better than the
other methods with more than 2\% absolute improvement. We note that
this comparison is not totally fair because we use images generated
from a generator pre-trained from ImageNet.  However, our contribution
is not to establish a new state-of-the-art score but to present the idea
of transferring an ImageNet pre-trained GAN for improving one shot
classifiers, so we believe this comparison is still informative.

\subsection{Results on NAB Dataset}
We also 
performed similar experiments on the NAB dataset, which is more than four times
larger than CUB, and the results are shown in the last column of
Table~\ref{tbl:results-main}. We observe  similar results as CUB,
 and that our method improves  classification accuracy
from a ProtoNet baseline (89.19\% vs. 87.91\%).

\section{Conclusion}\label{sec:conclusion}
We introduce an effective way to employ an ImageNet-pre-trained image
generator for the purpose of improving fine-grained one-shot
classification when data is scarce. As a way to fine-tune the
pre-trained generator, our pilot study finds that adjusting only scale
and shift parameters in batch normalization can produce a visually
realistic images. This technique works with a single image, making the method
less dependent on the number of available images. Furthermore,
although naively adding the generated images into the training set
does not improve  performance, we show that it can improve 
performance if we  mix generated with original
images to create hybrid training exemplars. In order to learn the parameters of this mixing, we adapt a
meta-learning framework.
 We implement this idea and demonstrate a consistent and significant
improvement over several classifiers on two fine-grained benchmark
datasets.
	
\subsubsection*{Acknowledgments}
We would like to thank Yi Li for drawing Figure 2, and Minjun Li and
Atsuhiro Noguchi for helpful discussions. Part of this work was done while
Satoshi Tsutsui was an intern at Fudan University. Yanwei Fu was supported in
part by the NSFC project (\#61572138), and Science and Technology
Commission of Shanghai Municipality Project (\#19511120700). 
David Crandall was supported in part by 
the National Science Foundation (CAREER IIS-1253549),
and the Indiana University
Office of the Vice Provost for Research, the College of Arts and
Sciences, and the Luddy School of Informatics, Computing, and Engineering
through the Emerging Areas of Research Project ``Learning: Brains,              
Machines, and Children.''
Yanwei Fu
is the corresponding author.

\clearpage

\bibliographystyle{plain}
\bibliography{references}

\clearpage

\section{Supplementary}
\subsection{Five-shot Experiments}
Although our paper focuses on one-shot learning, we also try a five-shot
scenario.  We use the ImageNet-pretrained ResNet18~\cite{he2016deep} as
a backbone. We also try the four layer-CNN (Conv-4) without ImageNet
pretraining in order to compare with other reported scores in a
benchmark study~\cite{chen2019closer}. The results are summarized in
Table \ref{tbl:results-cub-5shot}. When we use ImageNet pretrained
ResNet, our method is slightly (92.66\% v.s 92.83\%) better than the
ProtoNet. Given that non-meta-learning linear classifiers (softmax
regression and logistic regression) can achieve more than 92\%
accuracy, we believe that ImageNet features are already strong enough
when used with five examples per class.

\begin{table}[hbt]
	\caption{5-way-5-shot Accuracy (\%) on CUB dataset.}
	\label{tbl:results-cub-5shot}
	\centering
	\begin{tabular}{lllc}
		\toprule
	     Method           & Base Network    & Initialization & Accuracy   \\
	     \midrule
		Nearest neighbor     & ResNet18 & ImageNet & $89.44 \pm 0.36$        \\ 
		Softmax regression  & ResNet18 & ImageNet & $92.28 \pm 0.30$        \\ 
		Logistic regression   & ResNet18 & ImageNet & $92.34 \pm 0.30$        \\ 

         ProtoNet~\cite{snell2017prototypical} & ResNet18 & ImageNet & $92.97 \pm 0.31$        \\ 
         MetaIRNet (Ours) & ResNet18 & ImageNet  & $\mathbf{93.09 \pm 0.30}$       \\
         \midrule
 		 MAML\cite{finn2017model}            & Conv-4   & Random     & $72.09 \pm 0.76$~\cite{chen2019closer}         \\ 
 		 MatchingNet~\cite{vinyals2016matching} & Conv-4   & Random  & $72.86 \pm 0.76$~\cite{chen2019closer}  \\ 
 		 RelationNet\cite{sung2018learning}     & Conv-4   & Random  & $76.11 \pm 0.69$~\cite{chen2019closer}         \\ 
 		 ProtoNet~\cite{snell2017prototypical}  & Conv-4   & Random  & $80.75 \pm 0.46$         \\ 
 		 MetaIRNet (Ours)     & Conv-4   & Random  & $\mathbf{81.16 \pm 0.47}$         \\ 
		\bottomrule
	\end{tabular}
\end{table}

\subsection{An Implementation Detail: Class label input of BigGAN} Part of the noise $z$ used in BigGAN is class conditional, and we did not explicitly discuss this part in the main paper, so here we provide details.   We optimize the class
conditional embedding and regard it as part of the input noise.
Generally speaking, a conditional GAN uses input noise conditioned on
the label of the image to generate. BigGAN also follows this approach,
but our fine-tuning technique uses a single image to train. In other
words, we only have a single class label and can then optimize the
class embedding as part of the input noise.
 \vspace{0pt}

\subsection{More Experiments}
\paragraph{Increasing the number of generated examples.}  It is interesting to know if our method can benefit by increasing the number of generated examples. We try $n_{aug}=1, 2, 3, 5,10$ on CUB and obtain accuracies
 $84.13 \pm 0.60$, $83.45 \pm 0.60$, $80.99 \pm 0.62$, $81.21 \pm
0.62$, and $80.49 \pm 0.69$, respectively. Too many augmented images
seems to bias the classifier. We conclude that the performance gain is
marginal or even harmful when increasing $n_{aug}$.
 
\paragraph{Mixup baseline.} Mixup~\cite{zhang2018mixup}  uses random $1\times 1$ weights to mix two images, which can be viewed as a much simpler version of our method to mix a real and generated image pairs.  We test this
baseline and obtain one-shot accuracy of $82.24 \pm 0.59$ and $88.33
\pm 0.53$ on CUB and NAB, respectively. These results are higher than
baselines but still lower than ours.

\paragraph{Image deformation baseline.} Image deformation net~\cite{img_deform_2019} also uses similar $3\times3$ patch based data augmentation learning. The key difference is that while that method augments
support image by fusing with external real images called a gallery set,
our model fuses with images synthesized by GANs. Further, to adapt a generic
pretrained GAN to a new domain, we introduce a technique of optimizing
only the noise z and BatchNorm parameters rather than the full generator, which
is not explored by deformation net~\cite{img_deform_2019}. We try this baseline by using a gallery set of random  images sampled from the meta-training set, and obtain 1-shot-5-way accuracies of $82.84 \pm 0.62$
on CUB and $88.42 \pm 0.59$ on NAB, which is higher than the
baselines but not as high as ours. 

\paragraph{Training from scratch or end-to-end.} It is an interesting direction to train the generator end-to-end and without 
ImageNet pretraining.  Theoretically, we can do end-to-end training of
all components, but in practice we are limited by our GPU memory,
which is not large enough to hold both our model and BigGAN. In order
to simulate the end-to-end and scratch training, we introduce two
constraints. 1) We simplified BigGAN with one-quarter the number of
channels and train from scratch so that we train the generator with
a relatively small meta-training set. 2) We do not propagate the
gradient from classifier to the generator so that we do not have to
put both models onto GPU. We apply our approach with a 
four-layer CNN backbone with random initialization and achieved an 1-shot-5-way
accuracy of $63.77 \pm 0.71$ on CUB.

\paragraph{Experiment on Mini-ImageNet}. Although our method is designed for fine-grained recognition, it is interesting to apply this to coarse-grained recognition. Because the public BigGAN model was trained on images including the meta-testing set of ImageNet, we cannot use it as-is.  Hence we train the simplified generator (see above paragraph) from scratch using the meta-training set only. Using a backbone of ResNet18, the 1-shot-5-way accuracy on Mini-ImageNet is $53.97 \pm 0.63$ and $55.01 \pm 0.62$ for ProtoNet and MetaIRNet, respectively. 

\subsection{Dataset Split for NAB}
We used the following dataset split for the North American Bird
(NAB)~\cite{van2015building} dataset.  \small{
\paragraph{Label IDs used for training set} \texttt{295,  297,  299,  314,  316,  318,  320,  322,  324,  326,  328,
        330,  332,  334,  336,  338,  340,  342,  344,  346,  348,  350,
        352,  354,  356,  358,  360,  362,  364,  366,  368,  370,  372,
        374,  376,  378,  380,  382,  393,  395,  397,  399,  401,  446,
        448,  450,  452,  454,  456,  458,  460,  462,  464,  466,  468,
        470,  472,  474,  476,  478,  480,  482,  484,  486,  488,  490,
        492,  494,  496,  498,  500,  502,  504,  506,  508,  510,  512,
        514,  516,  518,  520,  522,  524,  526,  528,  530,  532,  534,
        536,  538,  540,  542,  544,  546,  548,  550,  552,  554,  556,
        558,  560,  599,  601,  603,  605,  607,  609,  611,  613,  615,
        617,  619,  621,  623,  625,  627,  629,  631,  633,  635,  637,
        639,  641,  643,  645,  647,  649,  651,  653,  655,  657,  659,
        661,  663,  665,  667,  669,  671,  673,  675,  677,  679,  681,
        697,  699,  746,  748,  750,  752,  754,  756,  758,  760,  762,
        764,  766,  768,  770,  772,  774,  776,  778,  780,  782,  784,
        786,  788,  790,  792,  794,  796,  798,  800,  802,  804,  806,
        808,  810,  812,  814,  816,  818,  820,  822,  824,  826,  828,
        830,  832,  834,  836,  838,  840,  842,  844,  846,  848,  850,
        852,  854,  856,  858,  860,  862,  864,  866,  868,  870,  872,
        874,  876,  878,  880,  882,  884,  886,  888,  890,  892,  894,
        896,  898,  900,  902,  904,  906,  908,  910,  912,  914,  916,
        918,  920,  922,  924,  926,  928,  930,  932,  934,  936,  938,
        940,  942,  944,  946,  948,  950,  952,  954,  956,  958,  960,
        962,  964,  966,  968,  970,  972,  974,  976,  978,  980,  982,
        984,  986,  988,  990,  992,  994,  996,  998, 1000, 1002, 1004,
       1006, 1008, 1010}

\paragraph{Label IDs used for validation set} \texttt{298,  315,  319,  323,  327,  331,  335,  339,  343,  347,  351,
        355,  359,  363,  367,  371,  375,  379,  392,  396,  400,  447,
        451,  455,  459,  463,  467,  471,  475,  479,  483,  487,  491,
        495,  499,  503,  507,  511,  515,  519,  523,  527,  531,  535,
        539,  543,  547,  551,  555,  559,  600,  604,  608,  612,  616,
        620,  624,  628,  632,  636,  640,  644,  648,  652,  656,  660,
        664,  668,  672,  676,  680,  698,  747,  751,  755,  759,  763,
        767,  771,  775,  779,  783,  787,  791,  795,  799,  803,  807,
        811,  815,  819,  823,  827,  831,  835,  839,  843,  847,  851,
        855,  859,  863,  867,  871,  875,  879,  883,  887,  891,  895,
        899,  903,  907,  911,  915,  919,  923,  927,  931,  935,  939,
        943,  947,  951,  955,  959,  963,  967,  971,  975,  979,  983,
        987,  991,  995,  999, 1003, 1007}

\paragraph{Label IDs used for test set} \texttt{296,  313,  317,  321,  325,  329,  333,  337,  341,  345,  349,
        353,  357,  361,  365,  369,  373,  377,  381,  394,  398,  402,
        449,  453,  457,  461,  465,  469,  473,  477,  481,  485,  489,
        493,  497,  501,  505,  509,  513,  517,  521,  525,  529,  533,
        537,  541,  545,  549,  553,  557,  561,  602,  606,  610,  614,
        618,  622,  626,  630,  634,  638,  642,  646,  650,  654,  658,
        662,  666,  670,  674,  678,  696,  700,  749,  753,  757,  761,
        765,  769,  773,  777,  781,  785,  789,  793,  797,  801,  805,
        809,  813,  817,  821,  825,  829,  833,  837,  841,  845,  849,
        853,  857,  861,  865,  869,  873,  877,  881,  885,  889,  893,
        897,  901,  905,  909,  913,  917,  921,  925,  929,  933,  937,
        941,  945,  949,  953,  957,  961,  965,  969,  973,  977,  981,
        985,  989,  993,  997, 1001, 1005, 1009}
}
\end{document}